\documentclass[english,5p,twocolumn,times]{elsarticle}
\usepackage[T1]{fontenc}
\usepackage[utf8]{inputenc}
\usepackage{geometry}
\usepackage{array}
\usepackage{verbatim}
\usepackage{textcomp}
\usepackage{multirow}
\usepackage{graphicx}
\usepackage{rotating}

\makeatletter

\providecommand{\tabularnewline}{\\}

\usepackage{color}
\usepackage{graphics}
\usepackage{array}
\usepackage{graphicx}
\usepackage{lastpage}
\usepackage{float}

\usepackage{times}

\usepackage{lineno}

\journal{XXX}

\usepackage{amsmath,amssymb}

\usepackage{colortbl}
\usepackage{multicol}

\definecolor{v1}{rgb}{0.0,0.8,0.0}
\definecolor{v2}{rgb}{0.8,1.0,0.0}
\definecolor{v3}{rgb}{0.8,0.8,0.8}
\definecolor{v4}{rgb}{0.9,0.5,0.0}
\definecolor{v5}{rgb}{0.8,0.0,0.0}

\makeatother

\usepackage{babel}
\begin{document}
\sloppy
\begin{frontmatter}

\title{A randomized simulation trial evaluating ABiMed, a clinical decision
support system\\
for medication reviews and polypharmacy management}

\author[label1]{Abdelmalek Mouazer}
\ead{malikmouazer@gmail.com}

\author[label5]{Sophie Dubois}
\ead{sophie.dubois@polesante13.fr}

\author[label1,label2]{Romain Léguillon}
\ead{romain.leguillon@chu-rouen.fr}

\author[label1]{Nada Boudegzdame}
\ead{nadaboudegzdame@gmail.com}

\author[label3]{Thibaud Levrard}
\ead{tlevrard@eig.fr}
\author[label3]{Yoann Le Bars}
\ead{ylebars@eig.fr}
\author[label3]{Christian Simon}
\ead{csimon@eig.fr}

\author[label1]{Brigitte Séroussi}
\ead{brigitte.seroussi@aphp.fr}

\author[label1,label2]{Julien Grosjean}
\ead{julien.grosjean@chu-rouen.fr}
\author[label1,label2]{Romain Lelong}
\ead{romain.lelong@chu-rouen.fr}
\author[label1,label2]{Catherine Letord}
\ead{catherine.letord@chu-rouen.fr}
\author[label1,label2]{Stéfan Darmoni}
\ead{stefan.darmoni@chu-rouen.fr} 

\author[label1]{Karima Sedki}
\ead{sedkikarima@yahoo.fr}

\author[label1]{Pierre Meneton}
\ead{pierre.meneton@inserm.fr}

\author[label7,label8]{Rosy Tsopra}
\ead{rosytsopra@gmail.com}

\author[label5]{Hector Falcoff}
\ead{hector.falcoff@sfr.fr}

\author[label1]{Jean-Baptiste Lamy\corref{cor1}}
\ead{jean-baptiste.lamy@inserm.fr}
\cortext[cor1]{Corresponding author}

\address[label1]{INSERM, Université Sorbonne Paris Nord, Sorbonne Université, Laboratory of Medical Informatics and Knowledge Engineering in e-Health, LIMICS, Paris, France}
\address[label2]{Department of Biomedical Informatics, Rouen University Hospital, France}
\address[label3]{EIG SAS, 92400 Courbevoie, France}
\address[label4]{Département de Médecine Générale, Université de Rouen, France}
\address[label5]{SFTG Recherche (Société de Formation Thérapeutique du Généraliste), Paris, France}
%\address[label6]{}
\address[label7]{Université Paris Cité, Sorbonne Université, Inserm, Centre de Recherche des Cordeliers, F-75006 Paris}
\address[label8]{Department of Medical Informatics, AP-HP, Hôpital Européen Georges-Pompidou, F-75015 Paris, France}
\begin{abstract}
\noindent \textbf{Background:} Medication review is a structured interview
of the patient, performed by the pharmacist and aimed at optimizing
drug treatments. In practice, medication review is a long and cognitively-demanding
task that requires specific knowledge. Clinical practice guidelines
have been proposed, but their application is tedious.

\noindent \textbf{Methods:} We designed ABiMed, a clinical decision
support system for medication reviews, based on the implementation
of the STOPP/START v2 guidelines and on the visual presentation of
aggregated drug knowledge using tables, graphs and flower glyphs.
We evaluated ABiMed with 39 community pharmacists during a randomized
simulation trial, each pharmacist performing a medication review for
two fictitious patients without ABiMed, and two others with ABiMed.
We recorded the problems identified by the pharmacists, the interventions
proposed, the response time, the perceived usability and the comments.
Pharmacists' medication reviews were compared to an expert-designed
gold standard.

\noindent \textbf{Results:} With ABiMed, pharmacists found 1.6 times
more relevant drug-related problems during the medication review ($p=1.1\times10^{-12}$)
and proposed better interventions ($p=9.8\times10^{-9}$), without
needing more time ($p=0.56$). The System Usability Scale score is
82.7, which is ranked ``excellent''. In their comments, pharmacists
appreciated the visual aspect of ABiMed and its ability to compare
the current treatment with the proposed one. A multifactor analysis
showed no difference in the support offered by ABiMed according to
the pharmacist's age or sex, in terms of percentage of problems identified
or quality of the proposed interventions.

\noindent \textbf{Conclusions:} The use of an intelligent and visual
clinical decision support system can help pharmacists when they perform
medication reviews. Our main perspective is the validation of the
system in clinical conditions.
\end{abstract}
\end{frontmatter}

\section{Introduction}

The worldwide population of people aged 65 or over is expected to
double, rising from 761 million to 1.6 billion by 2050 \citep{United2023}.
Many of them are exposed to polypharmacy (taking 5+ long term drugs
\citep{Gnjidic2012,Monegat2014}), which is associated with potentially
inappropriate medications (PIMs) and drug-related problem (DRP) including
adverse drug events (ADE) \citep{Lavan2016}. In France, inappropriate
prescriptions concern more than one in two elderly patients and their
direct cost is 507 million € per year \citep{Roux2022}.

General practitioners (GPs) are open to deprescribing inappropriate
drugs \citep{Jungo2021}, but often lack time and pharmacological
knowledge. Pharmacists could play a key role \citep{Croke2023}, by
conducting Medication Review (MR), “\emph{a structured evaluation
of a patient's medicines with the aim of optimizing medicines use
and improving health outcomes. This entails detecting drug-related
problems and recommending interventions}” \citep{Griese-Mammen2018}.
MR can change practices \citep{Zermansky2001}, reduce hospitalizations
\citep{Payen2022,Tasai2021} and saves costs \citep{Malet-Larrea2017,Garcia-Caballero2018}.
However, MR is a complex task, requiring efficient collaboration with
the GP, but also strong skills in clinical pharmacy \citep{Robberechts2024,Beuscart2021}.

To perform MR, pharmacists can be assisted by drug databases offering
a pharmaceutical description of all marketed drugs, and clinical practice
guidelines supporting MR \citep{Michiels-Corsten2020,Mouazer2021},
\emph{e.g.} STOPP/START \citep{O'Mahony2015,O'Mahony2023}. It includes
STOPP rules for detecting PIMs, and START rules for detecting potential
omissions. Studies demonstrated its ability to improve prescribing
appropriateness \citep{Gallagher2011}. However, drug databases provide
one page per drug, which is tedious when the patient takes many drugs,
and guidelines, in their paper format, are time-consuming and difficult
to apply in clinical routine \citep{Kurczewska-Michalak2021}.

\begin{figure*}
\includegraphics[viewport=0bp 0bp 601bp 272bp,clip,width=1\columnwidth]{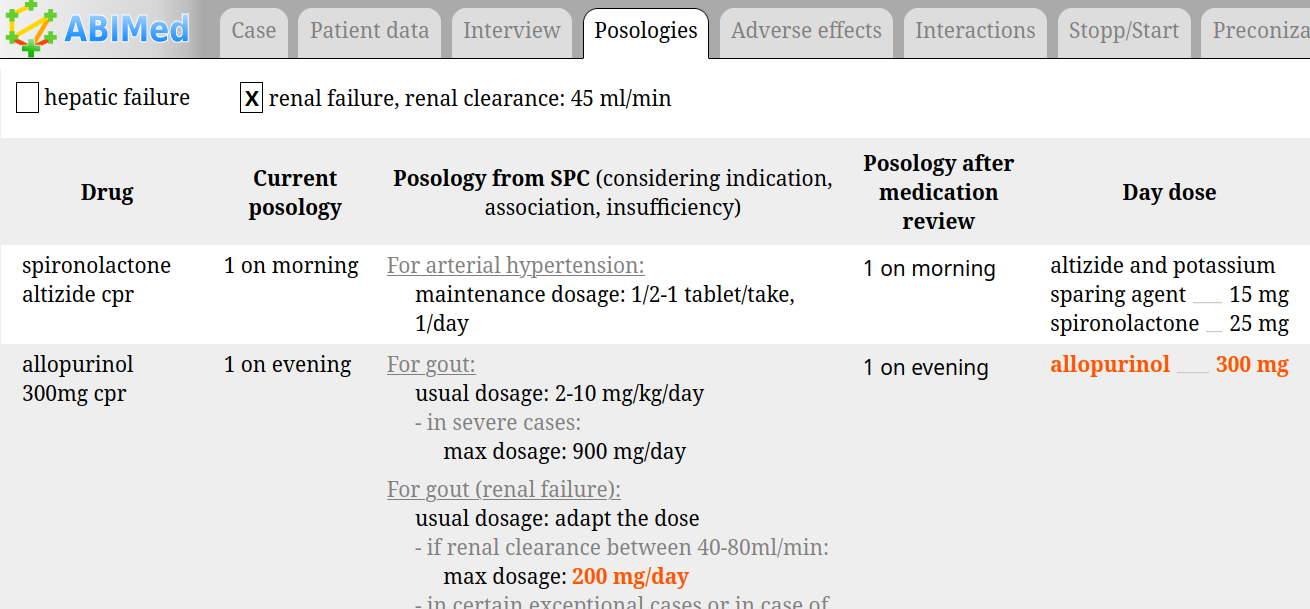}\hspace*{\fill}\includegraphics[clip,width=1\columnwidth]{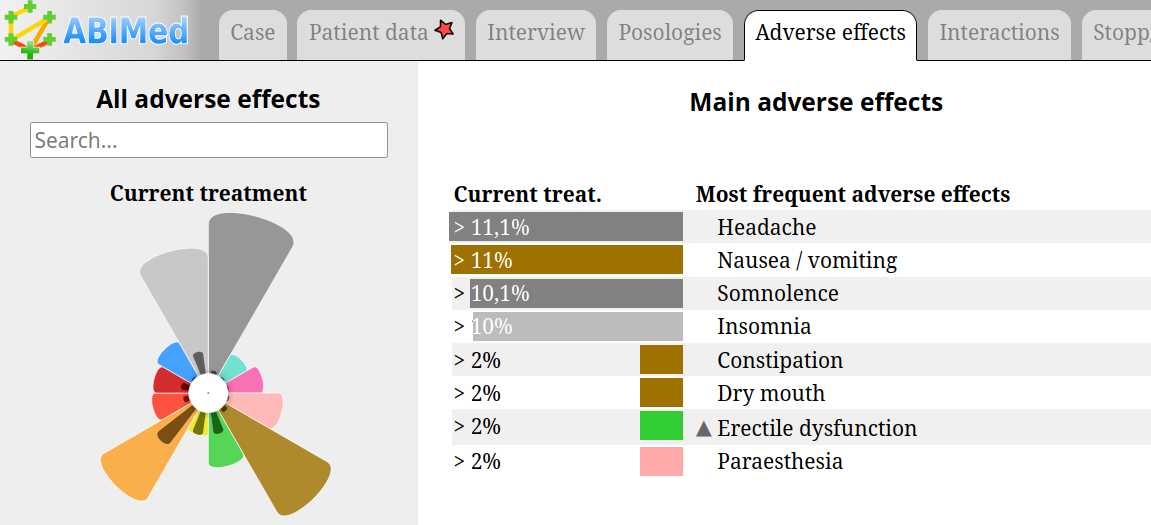}

\hspace*{\fill}A: dosage tab\hspace*{\fill}\hspace*{\fill}B: adverse
effect tab\hspace*{\fill}

~

\includegraphics[clip,width=1\columnwidth]{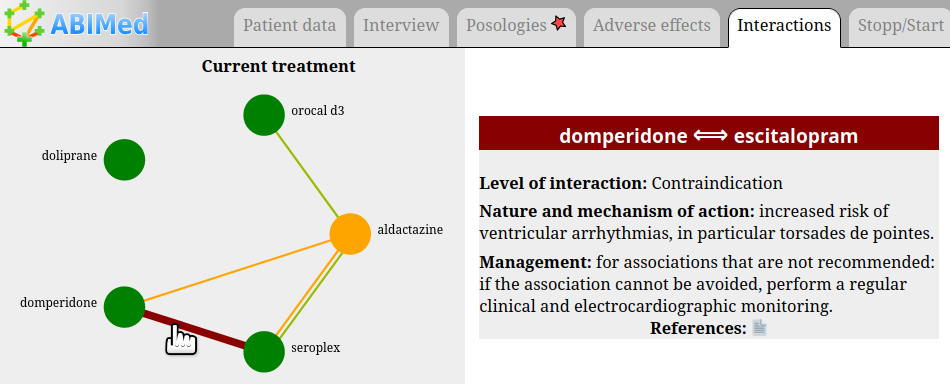}\hspace*{\fill}\includegraphics[clip,width=1\columnwidth]{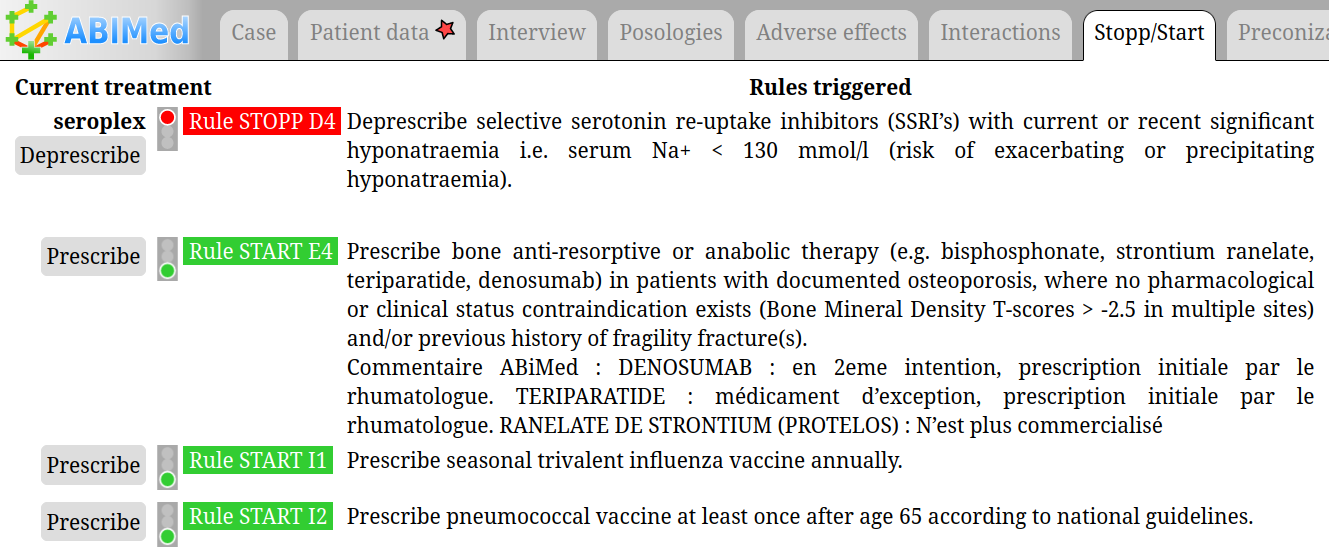}

\hspace*{\fill}C: drug interaction tab\hspace*{\fill}\hspace*{\fill}D:
STOPP/START tab\hspace*{\fill}

\caption{\label{fig:Screenshots}Screenshots of ABiMed's tabs for decision
support. A: table summarizing current dosages and recommended ones,
taking into account age, renal and hepatic status, indications and
drug associations; B: adverse effects profile expected with the entire
drug order, presented as flower glyphs; C: drug interactions, presented
as a radial graph; D: STOPP/START rules detected.}
\end{figure*}

Clinical Decision Support Systems (CDSSs) have been designed to overcome
these limitations \citep{Crowley2020}. A recent scoping review \citep{Mouazer2022_revue}
found 19 CDSSs for managing polypharmacy in the elderly. Most of them
implement paper guidelines by automating their recommendations, or
give access to drug knowledge. But the review highlighted limits:
the targeted user is often the physician instead of the pharmacist,
and the output is often displayed in a non-friendly format (\emph{e.g.}
long text).

To overcome these limitations, we designed ABiMed \citep{Mouazer2021_2,Mouazer2023_2},
a CDSS for helping community pharmacists to perform type-3 MR (\emph{i.e.}
including patient interview and clinical data \citep{Griese-Mammen2018}).
ABiMed associates two approaches: it implements STOPP/START rules
but also provides aggregated drug knowledge in a visual format, based
on visual analytics. The objective of this paper is to describe a
randomized simulation trial \citep{Cheng2016} evaluating ABiMed in
comparison to the usual practice. Our main hypothesis is that, with
ABiMed, pharmacists will be able to identify more PIMs and DRP, and
to propose better pharmaceutical interventions to solve these problems.

\section{\label{sec:Methods}Materials and methods}

\subsection{Brief description of ABiMed}

ABiMed is a CDSS for medication review and polypharmacy management
\citep{Mouazer2023_2,Mouazer2021_2}. It helps at three levels: (1)
it facilitates communication between pharmacists and GPs, and extracts
patient data from the GP's EHR; (2) it provides drug knowledge contextualized
for the patient, in a visual format (Figure \ref{fig:Screenshots});
and (3) it implements the STOPP/START v2 guidelines \citep{O'Mahony2015}
through a rule-based system that takes into account drugs (ATC classes
but also indications and doses), clinical conditions (ICD10 codes)
and biology (LOINC codes and associated values). ABiMed integrates
the Theriaque drug database \citep{Husson2008}. It proposes a comparative
mode, comparing the current treatment with the treatment after the
interventions proposed during MR, to check that these interventions
solve the problems identified without introducing new problems.

The current evaluation does not focus on the pharmacist-GP collaboration.
For more details on ABiMed, system design, ergonomic assessment and
qualitative evaluations, please refer to \citep{Mouazer2023_2}.

\subsection{Recruitment}

We recruited community pharmacists having a clinical practice in France.
We excluded pharmacists refusing to participate or without Internet
access. Pharmacists were recruited by emails, \emph{via} lists of
pharmacists from associations and territorial professional health
communities. They were compensated for their participation. Pharmacists
followed an online training session \emph{via} webinar (one hour).
We presented to them the project and its goal, the ABiMed software
and the evaluation principles.

\subsection{Clinical cases and gold standard}

An expert comity including two pharmacists (S Dubois, R Léguillon),
a GP (HF) and three researchers in health informatics with MD or PharmD
(AM, RT, JBL) designed four clinical cases, A, B, C, D, based on realistic
situations (Table \ref{tab:Metrics-of-the_case}). Each consisted
of an elderly patient with polypharmacy, consulting his/her pharmacist
for a problem leading to a MR. All clinical data was filled in the
case; thus, during MRs, no data entry was needed.

For each case, a gold standard was devised. It consisted of a list
of drug-related problems and, for each problem, a pharmaceutical intervention.
Interventions consisted of deprescribing a drug, prescribing a new
drug, replacing a drug, changing the dose, or prescribing lab tests.
Each intervention was assigned a score using CLEO v3 clinical scale
(CLinical, Economic and Organizational impacts of pharmacists' interventions)
\citep{Vo2021}, from 1 (minor) to 4 (vital).

\begin{table}
\begin{centering}
\begin{tabular}{ccccc}
Case & \#drugs & \#conditions & \#problem & total CLEO score\tabularnewline
\hline 
A & 8 & 8 & 6 & 10\tabularnewline
B & 7 & 7 & 5 & 7\tabularnewline
C & 5 & 6 & 6 & 7\tabularnewline
D & 5 & 5 & 5 & 11\tabularnewline
\hline 
\end{tabular}
\par\end{centering}
\caption{\label{tab:Metrics-of-the_case}Metrics of the four clinical cases.}
\end{table}

\begin{table}
\begin{centering}
\begin{tabular}{c>{\centering}p{1cm}>{\centering}p{1cm}>{\centering}p{1cm}>{\centering}p{1cm}}
\hline 
Intervention & \multicolumn{2}{c}{without ABiMed} & \multicolumn{2}{c}{with ABiMed}\tabularnewline
\hline 
Order of passage & \#1 & \#2 & \#3 & \#4\tabularnewline
\hline 
Group G1 & case A & case B & case C & case D\tabularnewline
Group G2 & case C & case D & case A & case B\tabularnewline
Group G3 & case B & case A & case D & case C\tabularnewline
Group G4 & case D & case C & case B & case A\tabularnewline
\hline 
\end{tabular}
\par\end{centering}
\caption{\label{tab:Groups}Definition of the four randomization groups.}
\end{table}

\subsection{Protocol}

The study followed a sequential group protocol, in which each pharmacist
first carried out two cases without ABiMed (\emph{i.e.} control cases)
online, and then two with ABiMed (\emph{i.e.} test cases). Pharmacists
were asked to use their usual resources when not having ABiMed. All
pharmacists performed MRs both with and without ABiMed and solved
the same cases, but with different order of passages. Pharmacists
were randomized into 4 groups (Table \ref{tab:Groups}). Control cases
were carried out before test cases, to avoid any learning phenomenon
when using ABiMed.

This protocol permits all pharmacists to test ABiMed, which allows
collecting the qualitative opinion of all pharmacists, and increases
the statistical power, each participant being its own control. Moreover,
in previous studies, we found that participants in the control group
lacked motivation and were more likely to give up, compromising the
entire study.

The INSERM ethics evaluation committee (IRB00003888) reviewed and
approved the study.

\subsection{\label{subsec:Data-collected}Data collected}

For each pharmacist, we collected:

(1) demographic data (online questionnaire): age group, gender (male,
female, other), number of MRs carried out the last year, previous
knowledge of STOPP/START (Boolean).

(2) for each case (automatic collection): problems identified and
pharmaceutical interventions proposed by the pharmacist, response
time.

(3) satisfaction questionnaire: SUS scale (System Usability Scale
\citep{Brooke1996}) comprising 10 questions with 5 possible answers,
and free comments.

Clinical cases were corrected semi-automatically. Interventions were
rated using CLEO scores, the maximum value was the value defined in
the gold standard, and lower values were given for suboptimal interventions.
The minimum value was -1, for harmful interventions. Interventions
proposed by pharmacists were automatically matched with the gold standard.
In case of match, we considered the problem as identified and the
intervention as having its maximum value. Then, cases were manually
and blindly reviewed by S Dubois, HF and JBL. We verified problems
and interventions, taking into account textual comments.

\subsection{Criteria}

The primary criterion was (1) the percentage of problems identified
in the MR, obtained by dividing the number of problems the pharmacist
identified by the number of problems in the gold standard.

Secondary criteria were:

(2) the CLEO score ratio, obtained by dividing the sum of the pharmacist's
interventions CLEO scores by the sum of the gold standard interventions
CLEO scores,

(3) the overall time spent to perform a MR,

(4) the pharmacist perceived usability, measured by SUS.

For example, a pharmacist identified 2 problems in case \#A and proposed
2 interventions, of CLEO score 1 and 2; the percentage of problems
identified is 33.3\% (since case \#A has 6 problems) and the CLEO
score ratio is 30\% (since case \#A has a total of 10 CLEO score and
only $1+2=3$ were proposed).

\subsection{Evaluation website}

The evaluation website for the pharmacists was based on ABiMed. It
includes two interfaces for MRs. The ``with ABiMed'' interface included
8 tabs: (1) the clinical case presentation, (2) the patient data (drugs
taken, clinical conditions and lab tests), (3) the patient interview,
(4-7) the decision support tabs (see Figure \ref{fig:Screenshots}),
(8) the interface for the pharmacist to enter his interventions and
comments. The ``without ABiMed'' interface was similar but without
tabs \#4-7.

\subsection{Statistical analysis}

Statistical analysis was performed blindly using R, with a risk $\alpha=5\%$
and bilateral tests. The unit of analysis is the MR, \emph{i.e.} a
given clinical case solved by a given pharmacist.

Criteria \#1-3 were compared using Welch two-Sample t-test. Times
were logged before comparison, to reduce the impact of long durations.
Linear Mixed Models (LMM) were used for finer analysis. Two models
were tested for each criterion \#1-3: a simple model considering three
fixed factors, ABiMed, Case and Group, and a complex model considering
ABiMed, Case, Age class of the pharmacist, Sex, Previous knowledge
of STOPP/START and Number of MRs performed the last year, as well
as their interactions with ABiMed. For each, the random effect was
modeled according to the pharmacist ID. Type-III ANOVA was used for
computing \emph{p}-values.

The carryover effect was evaluated by comparing criteria \#1-3 between
the first case treated by pharmacists and the second (both without
ABiMed), and the third and the fourth (both with ABiMed), respectively.

Supplementary file \#1 and \#2 contain the datasets, and \#3 contains
the R sources.

\begin{table}[t]
\begin{centering}
\begin{tabular}{l>{\raggedright}p{1.1cm}lc}
\textbf{Characteristic} & \textbf{Type} & \multicolumn{2}{c}{\textbf{Modalities / Aggregation}}\tabularnewline
\hline 
 &  & Male & 17 (44\%)\tabularnewline
Sex & nominal & Female & 22 (56\%)\tabularnewline
 &  & Other & 0\tabularnewline
\hline 
 & \multirow{5}{1.1cm}{integer} & 20-29 & 6 (15\%)\tabularnewline
 &  & 30-39 & 13 (33\%)\tabularnewline
Age &  & 40-49 & 10 (26\%)\tabularnewline
 &  & 50-59 & 7 (18\%)\tabularnewline
 &  & 60+ & 3 (8\%)\tabularnewline
\hline 
Previous knowledge & \multirow{2}{1.1cm}{nominal} & Yes & 29 (74\%)\tabularnewline
of STOPP/START &  & No & 10 (26\%)\tabularnewline
\hline 
 &  & Min & 0 (for 19 pharm.)\tabularnewline
\# MR in the  & \multirow{2}{1.1cm}{integer} & Max & 50 (for 1 pharm.)\tabularnewline
last year &  & Mean & 4.4\tabularnewline
 &  & Median & 1\tabularnewline
\hline 
\multirow{4}{*}{Group} & \multirow{4}{1.1cm}{nominal} & G1 & 10\tabularnewline
 &  & G2 & 9\tabularnewline
 &  & G3 & 11\tabularnewline
 &  & G4 & 9\tabularnewline
\hline 
\end{tabular}
\par\end{centering}
\begin{centering}
\par\end{centering}
\caption{\label{tab:demograph}Demographic characteristics of the recruited
pharmacists.}
\end{table}

\section{\label{sec:Results}Results}

\subsection{Recruited participants}

We recruited 39 pharmacists in the study (Table \ref{tab:demograph}).

\subsection{Impact on problem identification}

The left part of table \ref{tab:perf} shows the percentage of problems
identified by pharmacists without and with ABiMed. The overall percentage
is 45.0\% without ABiMed \emph{vs} 71.9\% with ABiMed, the difference
is highly significant ($p=1.1\times10^{-12}$). This shows that ABiMed
had a strong positive impact: using ABiMed, pharmacists identified
1.6 times more problems.

The left part of table \ref{tab:model_facteur} shows the LMM analysis
with the simple and complex models. The simple model confirms the
impact of ABiMed, and shows that there is a significant effect of
clinical cases, \emph{i.e.} some cases are simpler and some other
more difficult. The randomization group has no significant effect.

The complex model tests several characteristics of pharmacists, and
their interaction with ABiMed. Three factors are significant: ABiMed,
cases and age. Figure \ref{fig:problem_age} shows that the percentage
of problems identified slightly decreases with age. This is expected
because, in France, only younger pharmacists were trained in MR during
their studies.

\subsection{Impact on proposed interventions}

The middle part of table \ref{tab:perf} shows the CLEO score ratio
of the interventions proposed by pharmacists, without and with ABiMed.
The difference is highly significant, showing that pharmacists proposed
clinically better pharmaceutical interventions with ABiMed.

The complex LLM model (Table \ref{tab:model_facteur}, middle part)
identified three significant factors: ABiMed, clinical case, age,
and one interaction between ABiMed and case. Effect of age is similar
to its effect on the percentage of problems identified (see above).
Figure \ref{fig:cleo_cas} analyses the case-ABiMed interaction. For
case A, B and C, the use of ABiMed increased the CLEO score ratio,
but not for case D.

\subsection{Response times}

The right part of table \ref{tab:perf} shows mean and median response
times, without and with ABiMed. There is no significant difference.
Thus, the use of ABiMed did not require additional time for pharmacists,
but did not allow gaining time either. However, as pharmacists identified
more problems with ABiMed, the per-problem response time is lower.

\begin{figure*}[tp]
\noindent\begin{minipage}[t]{1\columnwidth}%
\begin{center}
\includegraphics[width=1\columnwidth]{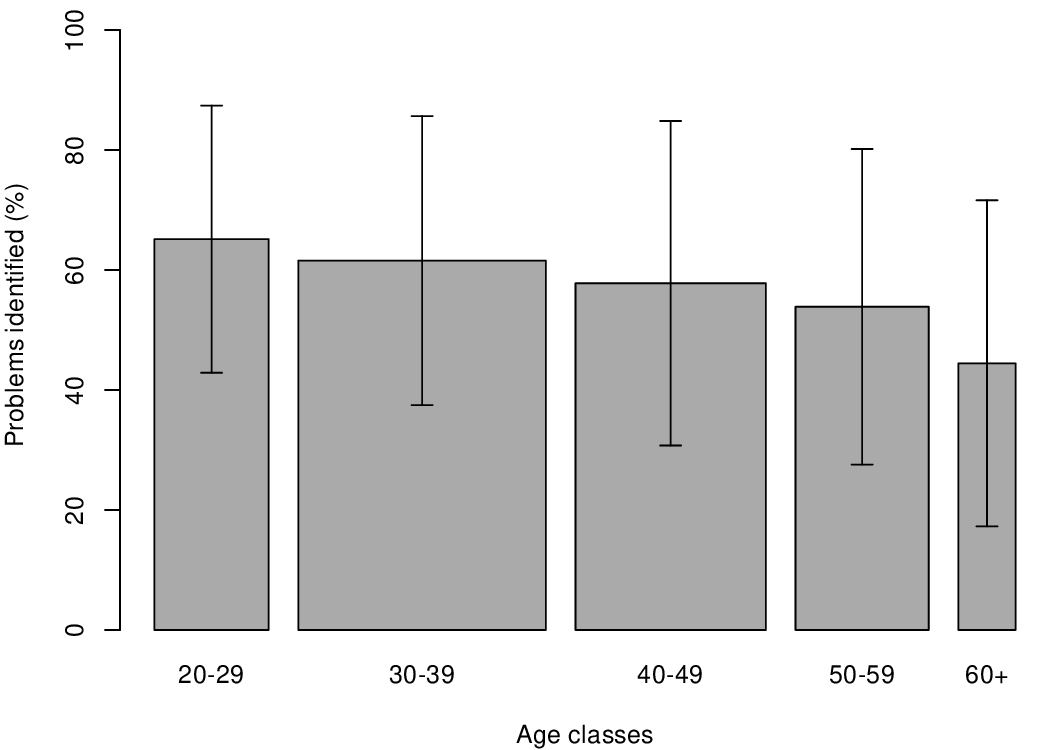}\vspace{-0.35cm}
\caption{\label{fig:problem_age}Bar plot showing the percentage of problems
identified, for each age class. Bar width is proportional to the number
of pharmacists.}
\par\end{center}%
\end{minipage}\hfill{}%
\noindent\begin{minipage}[t]{1\columnwidth}%
\begin{center}
\includegraphics[width=1\columnwidth]{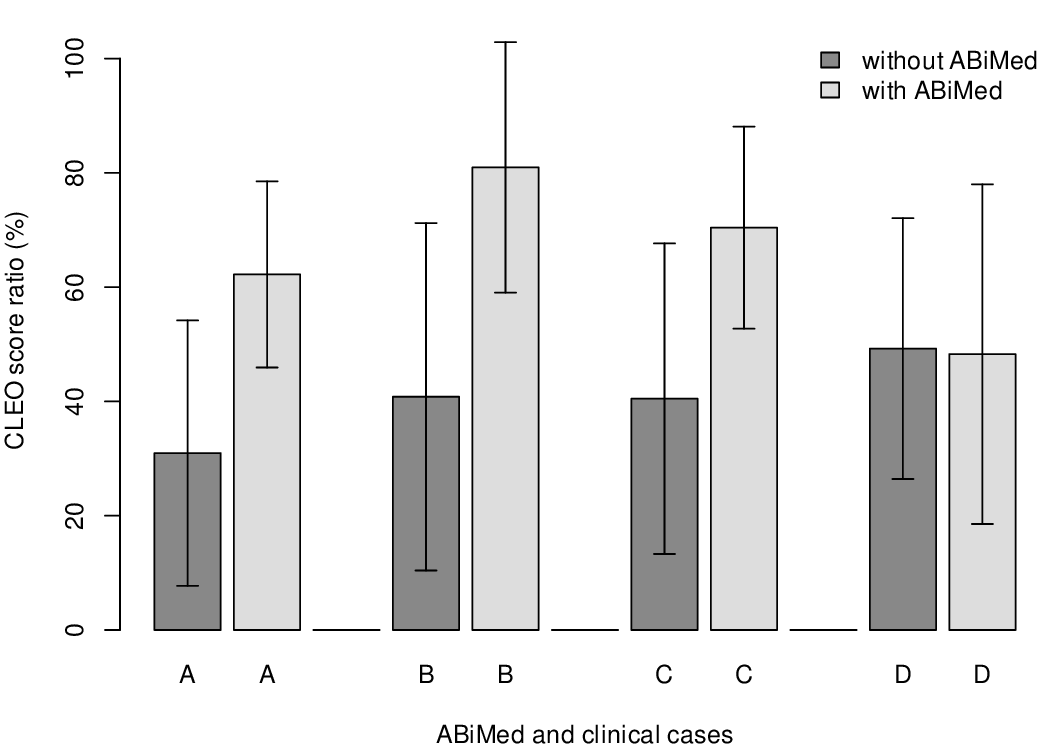}\vspace{-0.35cm}
\caption{\label{fig:cleo_cas}Bar plot showing the CLEO score ratio, without
or with ABiMed (left and right), for each clinical case (A-D).}
\par\end{center}%
\end{minipage}

\bigskip{}

\noindent\begin{minipage}[t]{1\columnwidth}%
\begin{center}
\includegraphics[width=1\columnwidth]{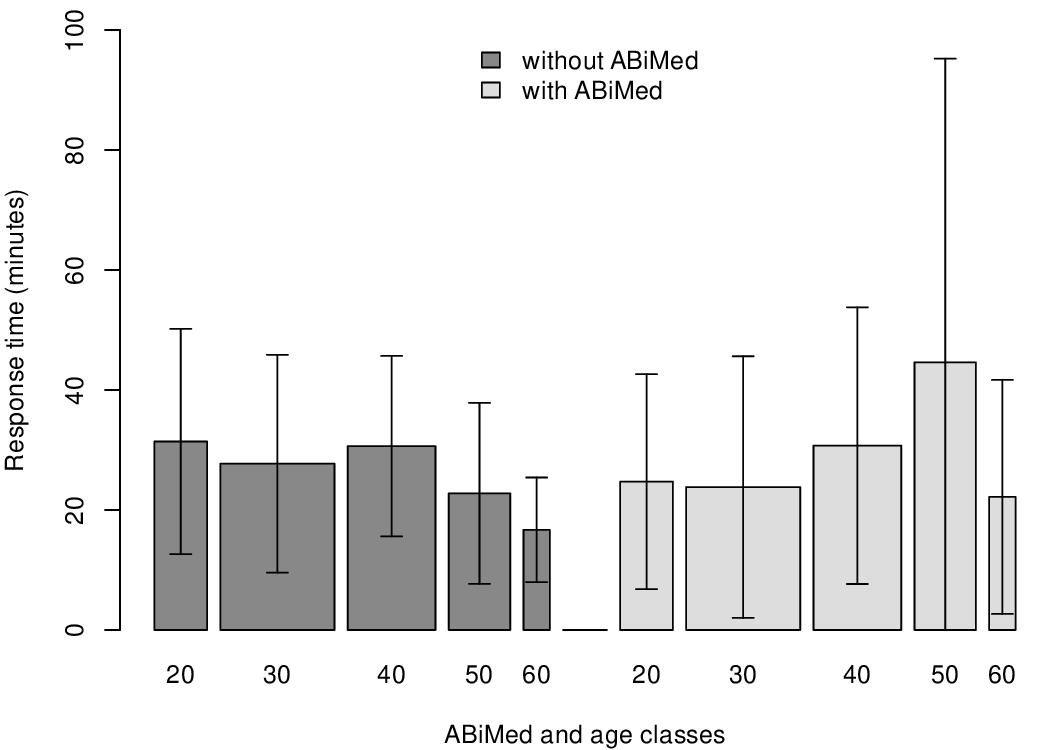}\vspace{-0.35cm}
\caption{\label{fig:time_age}Bar plot showing the response time, without or
with ABiMed, for each age class. Bar width is proportional to the
number of pharmacists.}
\par\end{center}%
\end{minipage}\hfill{}%
\noindent\begin{minipage}[t]{1\columnwidth}%
\begin{center}
\includegraphics[width=1\columnwidth]{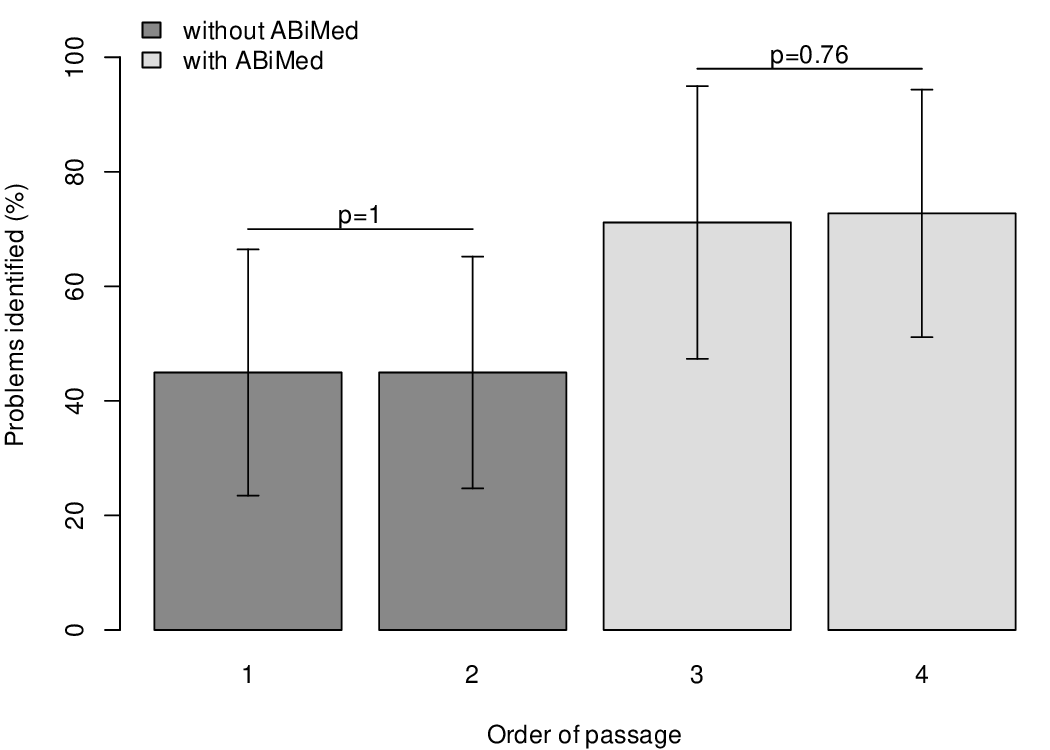}\vspace{-0.35cm}
\caption{\label{fig:carry-over}Bar plot showing the mean percentage of problems
identified according to the order of passage of the four clinical
cases.}
\par\end{center}%
\end{minipage}
\end{figure*}

The complex LLM model (Table \ref{tab:model_facteur}, right part)
identified one significant factor and one interaction: cases and ABiMed-age
interaction. Figure \ref{fig:time_age} suggests that, with ABiMed,
younger pharmacists were slightly faster and older pharmacists slightly
slower. Younger pharmacists might be more comfortable with computerized
CDSSs. However, the difference is limited and the size of subgroups
is small, so it should be interpreted cautiously.

\begin{table*}[p]
\begin{centering}
\begin{tabular}{c|cc|cc|cc}
\multicolumn{1}{c}{} & \multicolumn{2}{c}{\textbf{\% problems identified}} & \multicolumn{2}{c}{\textbf{interventions CLEO score ratio}} & \multicolumn{2}{c}{\textbf{mean / median time (minutes)}}\tabularnewline
\multicolumn{1}{c}{} & without ABiMed & \multicolumn{1}{c}{with ABiMed} & without ABiMed & \multicolumn{1}{c}{with ABiMed} & without ABiMed & with ABiMed\tabularnewline
\hline 
Case A & 46.8\% & 78.7\% & 31.0\% & 62.2\% & 30.5 / 28.3 & 35.0 / 21.9\tabularnewline
Case B & 43.8\% & 84.4\% & 40.8\% & 81.0\% & 22.9 / 24.7 & 41.4 / 25.1\tabularnewline
Case C & 43.5\% & 67.9\% & 40.5\% & 70.4\% & 25.1 / 24.1 & 16.0 / 15.1\tabularnewline
Case D & 45.6\% & 59.5\% & 49.2\% & 48.3\% & 31.0 / 22.5 & 27.4 / 23.9\tabularnewline
\hline 
\textbf{Overall} & \textbf{45.0\%} & \textbf{71.9\%} & \textbf{40.0\%} & \textbf{65.0\%} & \textbf{27.3 / 25.4} & \textbf{29.3 / 20.2}\tabularnewline
\emph{p}-value & \multicolumn{2}{c|}{$p=1.1\times10^{-12}$ {*}} & \multicolumn{2}{c|}{$p=9.8\times10^{-9}$ {*}} & \multicolumn{2}{c}{$p=0.56$}\tabularnewline
\hline 
\end{tabular}
\par\end{centering}
\caption{\label{tab:perf}Mean percentages of problems that have been identified
by pharmacists, mean interventions CLEO score ratios, and mean and
median response time, without or with ABiMed, for each clinical case,
and overall. {*}: difference is statistically significant when comparing
without \emph{vs} with ABiMed.}

\bigskip{}

\begin{centering}
\begin{tabular}{>{\raggedright}p{0.7cm}l|cc|cc|cc}
 & \multicolumn{1}{l}{} & \multicolumn{2}{c}{\textbf{\% problems identified}} & \multicolumn{2}{c}{\textbf{interventions CLEO score ratio}} & \multicolumn{2}{c}{\textbf{response time}}\tabularnewline
\textbf{Model} & \multicolumn{1}{l}{\textbf{Factor}} & \emph{p}-value & \multicolumn{1}{c}{inter. \emph{p}-value} & \emph{p}-value & \multicolumn{1}{c}{inter. \emph{p}-value} & \emph{p}-value & inter. \emph{p}-value\tabularnewline
\hline 
\multirow{3}{0.7cm}{\begin{turn}{90}
Simple
\end{turn}} & ABiMed & $<2.2\times10^{-16}$ {*} &  & $1.6\times10^{-12}$ {*} &  & 0.52~~~~ & \tabularnewline
 & Case & 0.0099 {*} &  & 0.015 {*} &  & 0.018 {*} & \tabularnewline
 & Group & 0.14~~~~ &  & 0.16~~~~ &  & 0.16~~~~ & \tabularnewline
\hline 
\multirow{6}{0.7cm}{\begin{turn}{90}
Complex
\end{turn}} & ABiMed & 0.0028 {*} & - & 0.0061 {*} & -~~~~ & 0.058~~~~ & -~~~~\tabularnewline
 & Case & 0.024 {*} & 0.31 & 0.00016 {*} & 0.0058 {*} (Fig \ref{fig:cleo_cas})\hspace{-0.5cm} & 0.022 {*} & 0.11~~~~\tabularnewline
 & Age class & 0.035 {*} (Fig \ref{fig:problem_age})\hspace{-1.1cm} & 0.79 & 0.044 {*} & 0.92~~~~ & 0.34~~~~ & 0.0098 {*} (Fig \ref{fig:time_age})\tabularnewline
 & Sex & 0.98~~~~ & 0.70 & 0.46~~~~ & 0.56~~~~ & 0.089~~~~ & 0.089~~~~\tabularnewline
 & STOPP/START known & 0.76~~~~ & 0.14 & 0.89~~~~ & 0.45~~~~ & 0.75~~~~ & 0.81~~~~\tabularnewline
 & \#MRs in the last year & 0.16~~~~ & 0.95 & 0.73~~~~ & 0.40~~~~ & 0.58~~~~ & 0.58~~~~\tabularnewline
\hline 
\end{tabular}
\par\end{centering}
\begin{comment}
\selectlanguage{french}%
\begin{center}
\begin{tabular}{l|cc|cc|cc}
\multicolumn{1}{l}{\selectlanguage{english}%
\selectlanguage{french}%
} & \multicolumn{2}{c}{\textbf{\% problèmes identifiés}} & \multicolumn{2}{c}{\textbf{interventions}} & \multicolumn{2}{c}{\textbf{temps de réponse}}\tabularnewline
\multicolumn{1}{l}{\textbf{Facteur}} & p & \multicolumn{1}{c}{inter. ABiMed} & p & \multicolumn{1}{c}{inter. ABiMed} & p & inter. ABiMed\tabularnewline
\hline 
ABiMed & \textbf{\textcolor{red}{0.0028 {*}}} & - & \textbf{\textcolor{red}{0.0061 {*}}} & -~~~~ & 0.058~~~~ & -~~~~\tabularnewline
Cas clinique & \textbf{\textcolor{red}{0.024 {*}}} & 0.31 & \textbf{\textcolor{red}{0.00016 {*}}} & \textbf{\textcolor{red}{0.0058 {*}}} & \textbf{\textcolor{red}{0.022 {*}}} & 0.11~~~~\tabularnewline
Âge & \textbf{\textcolor{red}{0.035 {*}}} & 0.79 & \textbf{\textcolor{red}{0.044 {*}}} & 0.92~~~~ & 0.34~~~~ & \textbf{\textcolor{red}{0.0098 {*}}}\tabularnewline
Sexe & 0.98~~~~ & 0.70 & 0.46~~~~ & 0.56~~~~ & 0.089~~~~ & 0.089~~~~\tabularnewline
Connaissance de STOPP/START & 0.76~~~~ & 0.14 & 0.89~~~~ & 0.45~~~~ & 0.75~~~~ & 0.81~~~~\tabularnewline
NB bilans par an & 0.16~~~~ & 0.95 & 0.73~~~~ & 0.40~~~~ & 0.58~~~~ & 0.58~~~~\tabularnewline
\hline 
\end{tabular}
\par\end{center}\selectlanguage{english}%
\end{comment}

\caption{\label{tab:model_facteur}\emph{p}-values obtained with Linear Mixed
Models (LMM) for each factor, and for their interactions with ABiMed
(\emph{i.e.} inter. \emph{p}-value) for the complex model, for each
of the three variables of interest (\% problems identified, CLEO score
ratio, response time). {*}: difference is statistically significant.}
\end{table*}

\begin{figure*}[t]
\begin{centering}
\includegraphics[width=0.89\textwidth]{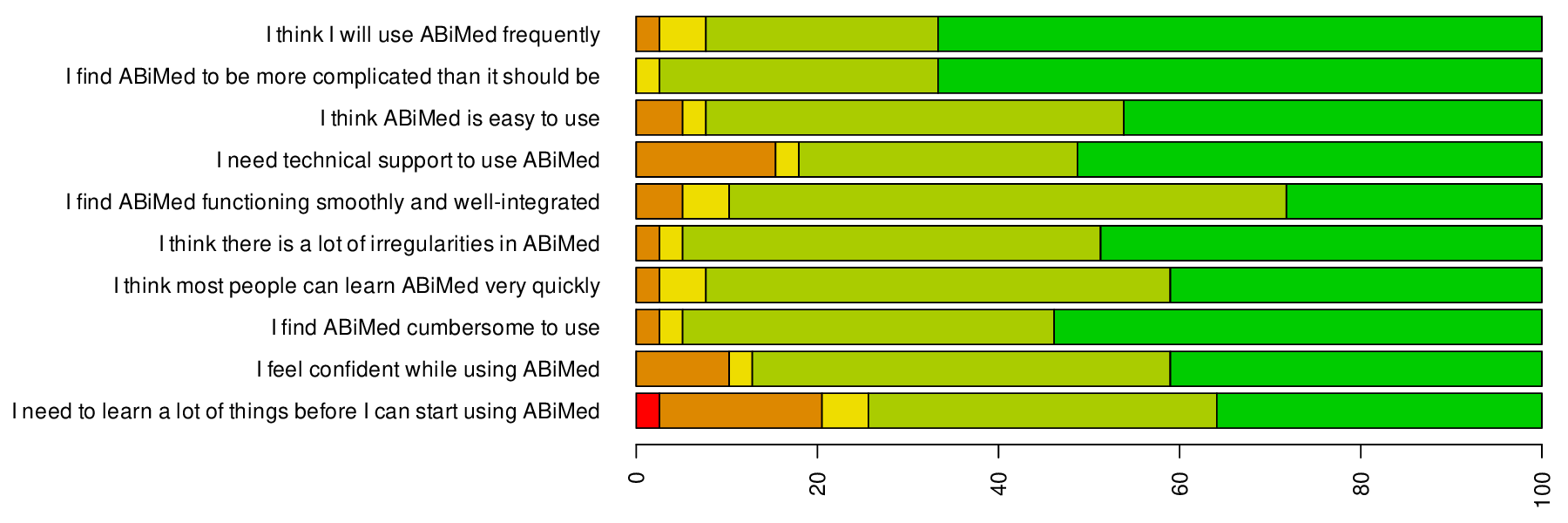}\includegraphics[width=0.11\textwidth]{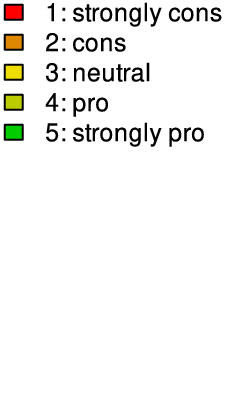}
\par\end{centering}
\caption{\label{fig:SUS}Distribution of answers to each question of the SUS
usability test, in percent (colors were reversed for even questions,
for which an approbation corresponds to a negative opinion about ABiMed:
here, ``pro'' and ``cons'' mean ``pro-ABiMed'' and ``cons-ABiMed'').}
\end{figure*}

\subsection{Carryover effect}

Figure \ref{fig:carry-over} shows the impact of the order of passage
of the four cases on the primary criteria. No significant differences
were found when comparing the percentage of problems identified between
cases \#1 and \#2 ($p=1$), and between cases \#3 and \#4 ($p=0.76$),
the CLEO score ratio ($p=0.37$ and $0.37$, respectively) and the
log(time) ($p=0.51$ and $0.18$, respectively). Thus, there was no
carryover effect. This was expected, the four cases covering distinct
clinical problems. On the contrary, the important difference between
cases \#1-2 and \#3-4 was caused by the intervention.

\subsection{Perceived usability}

The mean SUS score is 82.7 (55-100 depending on the pharmacists, Figure
\ref{fig:SUS}). This is ranked as ``excellent'' according to the
SUS adjective rating scale \citep{Bangor2009}. The less positive
answers were obtained for the last question, ``I need to learn a
lot of things before I can start using ABiMed''.

\begin{comment}
The LMM analysis found no significant difference in the SUS score
when testing the following factors: age class ($p=0.31$), sex ($p=0.57$),
number of MR performed in the last year ($p=0.14$) and previous knowledge
of STOPP/START ($p=0.27$). 
\end{comment}

\subsection{Pharmacist comments}

The pharmacist comments were enthusiastic, \emph{e.g.} ``the software
is really great, it is very reassuring'' or ``ABiMed greatly facilitates
the work of the pharmacist who performs a medication review and allows
him/her to quickly decide on the therapeutic decision to take.''

The pharmacists appreciated the visual aspects of ABiMed, including
tables (``the summary tables are very readable and allow us to have
an overview. Well done!''), the graph showing drug interactions (``I
find that the interaction tab is very practical'') and the flower
glyphs showing adverse effects (``the adverse effects tab is very
useful and visual: it could be used on a daily basis at the counter,
even outside of medication reviews! Well done''). They also appreciated
the comparative mode of ABiMed (``a good way to train and learn by
adding and testing several molecules together according to different
patient profiles.'').

Several pharmacists mentioned that ABiMed was very useful for identifying
problems, but more limited when it comes to proposing interventions
and suggesting replacement drugs. They proposed the implementation
of additional guidelines, targeting the main disorders for elderly
patients.

\section{\label{sec:Discussion}Discussion}

\begin{comment}
XXX rappel des résultats
\end{comment}

In this paper, we presented a simulation trial evaluating ABiMed,
a CDSS for supporting MR. Results show that, with ABiMed, pharmacists
identified more problems and proposed better interventions, without
spending more time.

\begin{comment}
XXX points forts
\end{comment}

The participants were community pharmacists with a clinical activity,
and not students, contrary to many studies. Recruiting professionals
is much more difficult, as they have less time available. As being
older, they may also be less open to innovation and digital tools.
However, they are more representative of the target users of ABiMed.

Results suggest that the support offered by ABiMed may vary according
to the patient profile: for problem identification, the mean is the
same for all cases without ABiMed (about 45\%) but differs with ABiMed
(60-85\%); for interventions (CLEO score ratio), no difference is
found for case D. This was related to an alert for domperidone that
was missing in ABiMed; the associated intervention (deprescription)
was ranked 4 on the CLEO scale. Actually, pharmacists performed better
with ABiMed for other problems of case D, but worse for the domperidone,
as they were relying on ABiMed guidance. This is known as \emph{automation
bias} \citep{Goddard2014}. Thus, we should take a particular care
to missing alerts.

Sex had no significant impact, as well as the sex-ABiMed interaction.
It means that the support provided by ABiMed is the same for males
and females, despite cognitive capacities sometimes differ, \emph{e.g.}
color perception \citep{Jameson2001}.

Surprisingly, the previous knowledge of STOPP/START and the number
of MRs performed in the last year had no significant impact. Possibly,
the knowledge of STOPP/START is not enough to execute the numerous
and complex rules manually, and many pharmacists did zero or one MR,
thus the impact of this factor may be difficult to analyze.

\begin{comment}
XXX limites
\end{comment}

The main limit of the study is that it is a simulation trial \citep{Cheng2016}:
pharmacists may not have acted as they would on real patients. However,
simulation trials are often used for CDSSs because of their simplicity
to set up, \emph{e.g.} \citep{Higi2023}. They also permit a better
comparison, since the same fictitious patients can be treated both
with and without the CDSS.

Another limit is that all patient data was available in the system,
thus pharmacists did not have to enter any additional data and just
focused on the analysis. This does not correspond to reality: pharmacists
often have limited access to patient data \citep{Nelson2017_2} (usually
stored in electronic health records owned by GPs or hospitals), and
some data is of questionable quality, or still in textual format,
that machines cannot interpret \citep{Kohane2021}. Data entry has
been a huge barrier when testing the PRIMA-eDS CDSS for managing polypharmacy
\citep{Rieckert2018}. To overcome this problem, we proposed in another
work an adaptive questionnaire for facilitating the entry of patient
clinical conditions for the automatic execution of STOPP/START rules
\citep{Lamy2023_2}. Nevertheless, the integration of ABiMed into
the medical workflow remains a matter of future research.

Finally, pharmacists were paid volunteers, which might bias the results.

\begin{comment}
XXX comparaison à la litérature
\end{comment}

In the literature, several CDSSs were clinically evaluated for the
support of MR, and were reviewed in \citep{Damoiseaux-Volman2022}.
However, most were less sophisticated than ABiMed, being limited to
triggering guideline-based alerts or detecting adverse reactions,
or focused on hospital rather than primary care. Hospital CDSSs for
polypharmacy obtain better results \citep{Scott2018}, possibly due
to the use of EHR. Outside hospital, examples include: the detection
of outlier prescriptions \emph{via} machine-learning for outpatients
\citep{Schiff2017}, the use of PHARAO, a system evaluating risk for
common side-effects in hospital and primary care \citep{Hedna2019},
the execution of STOPP rules in nursing homes \citep{Garcia-Caballero2018},
and the triggering of alerts from an electronic expert support system
in geriatric clinics and primary care \citep{Hammar2015}.

Contrary to ABiMed, few visual approaches were proposed, \emph{e.g.}
the use of colored tables to present interactions \citep{Schmidtchen2023},
and most systems targeted physicians rather than  pharmacists. Exceptions
include: a limited 5-rule alert system \citep{Mulder-Wildemors2020},
and a more substantial alert system but associated with a low sensitivity
\citep{Verdoorn2018}. In that context, the visual presentation of
drug knowledge can be a complementary approach, as it is less intrusive
than alerts, and not limited to a few situations but can include all
drug interactions.

\begin{comment}
XXX questions ouvertes et perspectives
\end{comment}

\section{\label{sec:Conclusion}Conclusion}

In conclusion, a simulation trial showed that ABiMed, a CDSS for helping
pharmacists perform MR, allowed them to identify 1.6 times more problems
than with their usual tools, and proposing better interventions, without
taking more time. Pharmacists appreciated ABiMed and gave it a high
perceived usability, ranked ``excellent''. Our main perspectives
are the improvement of ABiMed \emph{e.g.} integrating STOP/START v3
and clinical practice guidelines for common disorders to better support
the proposition of pharmaceutical interventions, the use of formal
argumentation to structure the discussion between pharmacists and
GPs, and the evaluation of ABiMed in clinical situations with real
patients, including the pharmacist-GP collaboration aspect.

\section{Summary table}

\subsection{What was already known on the topic}
\begin{itemize}
\item Medication review reduces hospitalizations and saves costs, but is
a long and tedious task for pharmacists.
\item Clinical practice guidelines for managing polypharmacy, such as STOPP/START,
are complex.
\item Most clinical decision support systems for polypharmacy focus on hospital,
physicians, and alert triggering.
\item On the contrary, ABiMed focuses on pharmacists, and combines the STOPP/START
v2 guidelines with visual drug knowledge using tables, graphs and
glyphs.
\end{itemize}

\subsection{What this study added to our knowledge}
\begin{itemize}
\item We conducted a randomized simulated trial comparing ABiMed to usual
resources used by pharmacists, on clinical cases.
\item Results show that, with ABiMed, pharmacists identify 1.6 times more
problems and proposed better intervention, without requiring more
time.
\item The perceived usability was ranked as ``excellent''.
\end{itemize}

\section*{CRediT authorship contribution statement}

Abdelmalek Mouazer: Conceptualization, Methodology, Software, Visualization

Sophie Dubois: Conceptualization, Methodology, Validation, Investigation,
Resources

Romain Léguillon: Methodology, Validation, Resources

Nada Boudegzdame: Methodology

Thibaud Levrard: Software

Yoann Le Bars: Methodology, Software

Christian Simon: Software, Supervision

Brigitte Séroussi: Methodology

Julien Grosjean: Methodology

Romain Lelong: Methodology

Catherine Letord: Methodology

Stéfan Darmoni: Supervision

Karima Sedki: Methodology

Pierre Meneton: Formal analysis, Writing - Review \& Editing

Rosy Tsopra: Methodology, Validation, Writing - Original Draft

Hector Falcoff: Conceptualization, Methodology, Validation, Investigation,
Resources

Jean-Baptiste Lamy: Conceptualization, Methodology, Software, Validation,
Formal analysis, Writing - Original Draft, Visualization, Supervision,
Project administration, Funding acquisition

\section*{Acknowledgments}

This work was funded by the French Research Agency (ANR) through the
ABiMed project {[}Grant No. \mbox{ANR-20-CE19-0017}{]}.

\bibliographystyle{elsarticle-num}
\bibliography{biblio_ama,biblio_ecmt}

\end{document}